%% file: step.tex
\documentclass{article}

\usepackage[nonatbib, final]{neurips_data_2021}

\usepackage[utf8]{inputenc} 
\usepackage[T1]{fontenc}    
\usepackage[pagebackref=true,breaklinks=true,letterpaper=true,colorlinks,bookmarks=false]{hyperref}       
\usepackage{url}            
\usepackage{booktabs}       
\usepackage{amsfonts}       
\usepackage{nicefrac}       
\usepackage{microtype}      

\usepackage{amsmath}
\usepackage{amssymb}
\usepackage{bm}
\usepackage[dvipsnames]{xcolor}  

\usepackage[pdftex]{graphicx}
\usepackage{caption}
\usepackage{subcaption}
\usepackage{cuted}

\usepackage{tikz}
\usepackage{enumitem}
\usepackage{pifont}
\newcommand{\cmark}{\ding{51}}
\newcommand{\xmark}{\ding{55}}

\newcommand{\cbox}[1]{\tikz[baseline=-0.5ex]\draw[#1, line width=3, ](0,0) -- (0.2, 0);}

\makeatletter
\DeclareRobustCommand\onedot{\futurelet\@let@token\@onedot}
\def\@onedot{\ifx\@let@token.\else.\null\fi\xspace}

\def\eg{\emph{e.g}\onedot} 
\def\ie{\emph{i.e}\onedot} 
\def\cf{\emph{c.f}\onedot} 
 \def\vs{\emph{vs}\onedot}
\def\wrt{w.r.t\onedot} 
\def\etal{\emph{et al}\onedot}
\makeatother

\usepackage{xspace}

\input{macro}
\input{abbrev}

\title{STEP: Segmenting and Tracking Every Pixel}

\author{%
  Mark Weber$^1$\thanks{This work was partially done during an internship. $^1$Technical University Munich, $^2$Google Research, $^3$RWTH Aachen University, $^4$MPI-IS and University of T{\"u}bingen}
  \And
  Jun Xie$^2$
  \And
  Maxwell Collins$^2$
  \And
  Yukun Zhu$^2$
  \And
  Paul Voigtlaender$^{2, 3}$
  \And
  Hartwig Adam$^2$
  \And
  Bradley Green$^2$
  \And 
  Andreas Geiger$^4$
  \And 
  Bastian Leibe$^3$
  \And 
  Daniel Cremers$^1$
  \And 
  Aljo{\v{s}}a O{\v{s}}ep$^1$
  \And
  Laura Leal-Taix{\'e}$^1$
  \And
  Liang-Chieh Chen$^2$
}

\begin{document}

\maketitle
\vspace{-15pt}
\begin{center}
 \includegraphics[width=\textwidth]{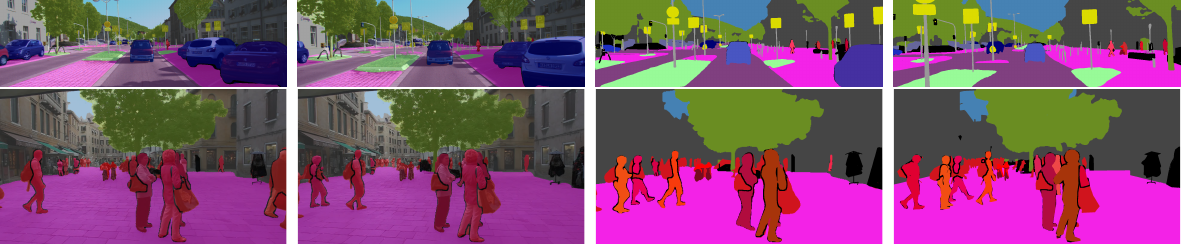}
 \vspace{-15pt}
 \captionof{figure}{Our proposed ground-truth labels of \kitti (top) and \challenge (bottom).
 }
 \label{fig:teaser}
 \vspace{0.2cm}
\end{center}%

\input{sections/0b.abstract}

\input{sections/1b.intro}
\input{sections/2.related}

\input{sections/3b.dataset-iccv}
\input{sections/4.metric}
\input{sections/5.baselines}
\input{sections/6.experiments}
\input{sections/7.conclusion}

{\small
\bibliographystyle{ieee_fullname}
\bibliography{egbib,references}
}

\input{sections/9.supplementary}


\end{document}

%% file: macro.tex
\newcommand{\refsec}[1]{Sec.\,\ref{sec:#1}}

\newcommand{\figref}[1]{Fig.\,\ref{fig:#1}}
\newcommand{\tabref}[1]{Tab.\,\ref{tab:#1}}

\newcommand{\PAR}[1]{\vskip4pt \noindent {\bf #1~}}

\newcommand{\changed}[1]{#1}

%% file: abbrev.tex
\newcommand*{\kitti}{{KITTI-STEP}\@\xspace}
\newcommand*{\challenge}{{MOTChallenge-STEP}\@\xspace}

\newcommand*{\metric}{{STQ}\@\xspace}
\newcommand*{\metricfull}{{Segmentation and Tracking Quality}\@\xspace}

\newcommand*{\tp}{TP_c} 
\newcommand*{\fn}{FN_c} 
\newcommand*{\fp}{FP_c} 
\newcommand*{\idsw}{IDSW_c}

%% file: sections/0b.abstract.tex
%

\begin{abstract}
  The task of assigning semantic classes and track identities to every pixel in a video is called video panoptic segmentation.
  Our work is the first that targets this task in a real-world setting requiring dense interpretation in both spatial \emph{and} temporal domains.
  As the ground-truth for this task is difficult and expensive to obtain, existing datasets are either constructed synthetically or only sparsely annotated within short video clips.
  To overcome this, we introduce a new benchmark encompassing two datasets, \kitti, and \challenge.
  The datasets contain long video sequences, providing challenging examples and a test-bed for studying long-term pixel-precise segmentation and tracking under real-world conditions.
  %
  %
  We further propose a novel evaluation metric \metricfull (\metric) that fairly balances semantic and tracking aspects of this task and is more appropriate for evaluating sequences of arbitrary length. 
  Finally, we provide several baselines to evaluate the status of existing methods on this new challenging dataset.
  We have made our datasets, metric, benchmark servers, and baselines publicly available, and hope this will inspire future research. 
\end{abstract}

%% file: sections/1b.intro.tex
\vspace{-10pt}
\section{\changed{Introduction}}


%
%
Dense and pixel-precise video scene understanding is of fundamental importance for applications such as autonomous driving, film editing and spatio-temporal reasoning. More specifically, while the semantic interpretation helps with tasks such as estimating the drivable area for an autonomous vehicle, the tracking of objects enables us to anticipate the temporal evolution of the surroundings, which is critical for motion planning and obstacle avoidance. 

\PAR{Challenges.} Moving towards this goal, there are three challenges that we find not addressed by previous benchmarks. First, the ability of \emph{explaining every pixel} of a continuous input from cameras.  Second, changes in the input signal over time can happen quickly and hence, demanding that \emph{we evaluate with the same high-frequency} as that of the changes that occur. Third, interpretation of continuous sensory input requires temporally consistent scene understanding, \ie, \emph{long-term tracking}, which current benchmarks and metrics are not suitable for.
The aim of this work is to advance this field by introducing a suitable benchmark and metric.

In the past, image benchmarks such as PASCAL VOC \cite{Everingham10IJCV}, ImageNet \cite{Russakovsky15IJCV}, and COCO \cite{Lin14ECCV} played a pivotal role in the astonishing progress of computer vision research over the last decade, allowing the community to evaluate different methods in a standardized way.
Real-world datasets for various tasks~\cite{Krizhevsky12NIPS, Szegedy15CVPR, Simonyan15ICLR, Shaoqing15NIPS, He17ICCV, Chen18ECCV}
were used to fairly measure the progress and highlight key innovations. 

For the purpose of holistic image understanding, Kirillov \etal~\cite{Kirillov19CVPR} introduced the concept of \textit{panoptic segmentation} as a combination of semantic segmentation and instance segmentation.
Kim \etal~\cite{Kim20CVPR} subsequently introduced the notion of video panoptic segmentation (VPS). 
Yet, they merely label a sparse subset of pixels from short real-world video snippets that are not suitable for dense pixel-precise video understanding. Moreover, existing \emph{synthetic} datasets~\cite{Hurtado20CVPRW, Kim20CVPR} struggle to evaluate performance in the real-world due to the domain shift~\cite{Sankaranarayanan2018CVPR}. 

For evaluation of VPS, existing metrics~\cite{Hurtado20CVPRW, Kim20CVPR} build upon metrics for panoptic segmentation and multi-object tracking. Since a metric can be significant in deciding the community's research direction, biases in the metric can hinder promising innovations.

\PAR{Contributions.} The contribution of this work is threefold:
%
%

(1) We introduce more suitable benchmark datasets that in particular allow spatio-temporally dense and pixel-centric evaluation. Our proposed benchmark extends the existing KITTI-MOTS, and MOTS-Challenge datasets \cite{Voigtlaender19CVPR_mots} with spatially and temporally dense annotations. 
We seek to label \emph{every} pixel with a semantic class and track ID. As in panoptic segmentation~\cite{Kirillov19CVPR}, we treat each non-countable region, such as the \textit{sky}, as belonging to a single track. For the most salient countable classes, each instance is assigned a semantic class and a unique ID throughout the video sequences.

(2) After studying prior metrics in detail in \refsec{metric}, we propose the \metricfull (\metric) metric that is more suitable to access the segmentation and tracking performance of algorithms. \metric is defined at the pixel level and provides an accurate and intuitive comparison against the ground-truth at a fine-grained level. The core principle of our benchmark is that each pixel in each frame matters when evaluating an algorithm. 
%
%

(3) Finally, our datasets and metric provide us a valid test-bed for evaluating several baselines that show the effect of unified \vs separate (\cf \refsec{baselines}) and motion- \vs appearance-based methods on our benchmark.
This includes methods that use optical flow for mask propagation~\cite{Osep18ICRA,Luiten18ACCV,Teed20ECCV} or methods inspired from state-of-the-art tracking work~\cite{Bergmann19ICCV,Zhou20ECCV}. 
Test servers will enable a fair benchmark of methods. 
This provides a complete framework to enable research into dense video understanding,
where both segmentation and tracking are evaluated in a detailed and holistic way. 
%
In summary,
\begin{itemize}[itemsep=2pt]
    \item We present the first real-world spatially and temporally dense annotated datasets \kitti and \challenge, providing challenging segmentation and (long) tracking scenes (\refsec{datasets}).
    \item We analyze in-depth the recently proposed metrics \cite{Kim20CVPR, Hurtado20CVPRW}, and based on our findings propose the \metricfull (\metric) metric (\refsec{metric}).
    \item We showcase simple baselines based on established segmentation and tracking paradigms, motivating future research in end-to-end models (\refsec{baselines}, \refsec{results}).
\end{itemize}

%% file: sections/2.related.tex
\section{Related Work}


\PAR{Panoptic Segmentation.} The task of panoptic segmentation combines semantic segmentation~\cite{He2004CVPR} and instance segmentation~\cite{Hariharan2014ECCV}, and requires assigning a class label and instance ID to all pixels in an image.
It is growing in popularity, thanks to the proposed benchmark~\cite{Kirillov19CVPR}. The corresponding panoptic quality (PQ) metric is the product of recognition quality and segmentation quality.


 
\PAR{Video Semantic Segmentation.} Compared to image semantic segmentation~\cite{Everingham10IJCV,Long2015CVPR,Chen2015ICLR}, video semantic segmentation~\cite{Zhu2017CVPR} is less common in the literature, possibly due to the lack of benchmarks. Video semantic segmentation differs from our setting in that it does not require discriminating different instances and hence, also no explicit tracking.
 
\PAR{Multi-Object Tracking.} The task of multi-object tracking (MOT) is to accurately track multiple objects by associating bounding boxes in videos~\cite{Breitenstein2009ICCV,Bergmann19ICCV,Peng2020ECCV}. Focusing on tracking, this task does not require any segmentation.
The MOT-Challenge~\cite{dendorfer2020ijcv,Dendorfer2020Arxiv} and KITTI-MOT \cite{Geiger12CVPR} datasets are among the most popular benchmarks, and the tracking performance is measured by the CLEAR MOT metrics~\cite{Bernardin08JIVP} along with a set of track quality measures introduced in~\cite{Wu2006CVPR}.
Recently, \cite{Luiten20IJCV} propose HOTA (Higher Order Tracking Accuracy), which explicitly balances tracking results attributed from accurate detection and association.


\PAR{Video Instance Segmentation.} Combining instance segmentation and MOT, the goal of video instance segmentation (VIS)~\cite{Yang19ICCV} is to track instance masks across video frames. 
This task is also known as multi-object tracking and segmentation (MOTS)~\cite{Voigtlaender19CVPR_mots}.
Yet, the respective work adapt different evaluation metrics, focusing on different perspectives of the task.
The prominent Youtube-VIS dataset~\cite{Yang19ICCV} focuses on short sequences (3-6 seconds, every 5th frame annotated) in which objects are present from start to end. For that, ${AP}^{mask}$~\cite{Hariharan2014ECCV,Lin14ECCV} is adapted to videos by extending the IoU to the temporal domain (3D IoU). The focus of MOTS datasets, KITTI-MOTS, and MOTS-Challenge~\cite{Voigtlaender19CVPR_mots}, is on more general scenarios with appearing and disappearing objects within long videos. Therefore, MOTSA (Multi-Object Tracking and Segmentation Accuracy) is used, the mask-based variants of the CLEAR MOT metrics~\cite{Bernardin08JIVP}.
The two MOTS datasets focus on challenging urban sequences like one encounters in an autonomous driving setting.
In contrast to our task, these benchmarks do not consider non-instance regions and hence, do not require pixel-precise video understanding. 
We build on top of these datasets by adding {\it semantic segmentation} annotations to obtain \kitti and \challenge.



\PAR{Video Panoptic Segmentation.} Recently, panoptic segmentation has also been extended to the video domain.
Video Panoptic Segmentation (VPS)~\cite{Kim20CVPR} requires generating the instance tracking IDs along with panoptic segmentation results across video frames.
Previous datasets~\cite{Kim20CVPR, Hurtado20CVPRW} and corresponding metrics fail to properly evaluate scenarios that require both short- and long-term segmentation and tracking.
Specifically, due to expensive annotation efforts for VPS datasets, most of the reported experimental results in the literature~\cite{Kim20CVPR,Hurtado20CVPRW} are conducted on synthetic datasets~\cite{Richter2017ICCV,Gaidon2016CVPR}, making it hard to generalize to real-world applications~\cite{Hoffman2018ICML, tsai2018CVPR}.
Exceptionally, \cite{Kim20CVPR} annotates Cityscapes video sequences~\cite{Cordts16CVPR} for 6 frames out of each short 30 frame clip (about 1.8 seconds), hence focusing on short-term tracks.
Both benchmarks~\cite{Kim20CVPR, Hurtado20CVPRW} evaluate with different metrics, VPQ (Video Panoptic Quality) and PTQ (Panoptic Tracking Quality), respectively.
VPQ, tailored for sparse annotations on short clips, was not designed for evaluating long-term tracking, while PTQ penalizes recovery from association errors and may produce hard to interpret negative evaluation results.
Motivated by the issues above, we propose two datasets densely annotated in space and time, \kitti and \challenge requiring {\it long-term} segmentation and tracking. Additionally, we propose a novel \metric metric that gives equal importance to segmentation and tracking.

%% file: sections/3b.dataset-iccv.tex
\section{Datasets}
\label{sec:datasets}

\begin{figure}[t]
    \centering
    \includegraphics[width=0.8\textwidth]{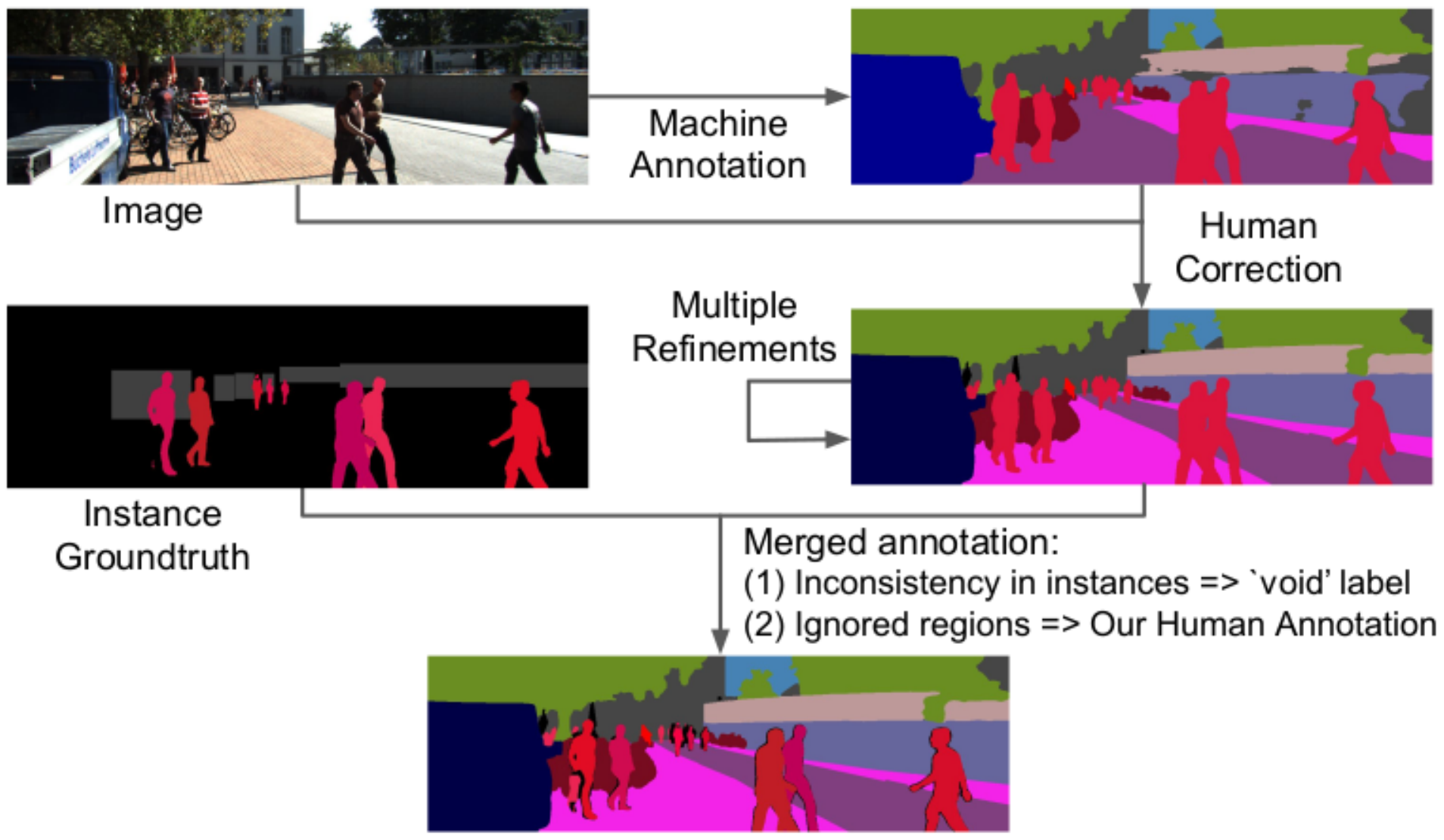}
     \caption{Annotation process: The machine annotation {\it semantic segmentation} from Panoptic-DeepLab is corrected by human annotators with multiple refinements. The resulting annotation is further merged with the existing instance ground-truth from KITTI-MOTS and MOTS-Challenge.
     }
     \label{fig:dataset_main}
\end{figure}

\PAR{Overview.}
We collect two densely annotated datasets in both spatial and temporal domains, \ie, every pixel in every frame is annotated, building on top of KITTI-MOTS and MOTS-Challenge~\cite{Voigtlaender19CVPR_mots}.
Both have carefully annotated tracking IDs for `pedestrians' and `cars' (KITTI-MOTS only) in real-world scenes, as they are the most salient moving objects. 
So far, other common semantic classes in urban scenes, such as `bicycles' and `road', have all been grouped into one `background' class, impeding pixel-level scene understanding.
Therefore, our additional {\it semantic segmentation} annotation, defined by the 19 Cityscapes classes~\cite{Cordts16CVPR},  enriches the KITTI-MOTS and MOTS-Challenge datasets. Specifically, all `background' regions are carefully annotated, and classes that are not tracked, \ie, everything except `pedestrians' and `cars' in our case, are considered as a single track, similar to the `stuff' definition in panoptic segmentation~\cite{Kirillov19CVPR}). The resulting datasets, called \kitti and \challenge, present challenging videos requiring long-term consistency in segmentation and tracking under real-world scenarios. \figref{teaser} shows ground-truth annotations from the proposed dataset.

\PAR{Semi-automatic annotation.} Similar to \cite{Castrejon2017CVPR,Andriluka2018ACMM,Voigtlaender19CVPR_mots}, we collect our annotations in a semi-automatic manner. In particular, we employ the state-of-the-art Panoptic-DeepLab~\cite{Cheng20CVPR,Chen20ECCV}, pretrained on the Mapillary Vistas~\cite{Neuhold2017ICCV} and Cityscapes~\cite{Cordts16CVPR} datasets to generate pseudo {\it semantic} labels for each frame. The predicted {\it semantic segmentation} is then carefully refined by human annotators.
On average, it takes human annotators around 10 minutes to annotate each of the more than 20000 frames. The refining process is iterated twice to guarantee high-quality per-frame annotations and consistent labels across consecutive video frames. We illustrate our semi-automatic annotation process in \figref{dataset_main} and provide all technical details regarding the merging procedure in the supplement. It is evident from the illustration, that human refinements and corrections are required to achieve high quality annotations.

\begin{figure}[t]
    \centering
    \begin{tabular}{c c}
    \includegraphics[width=0.49\textwidth]{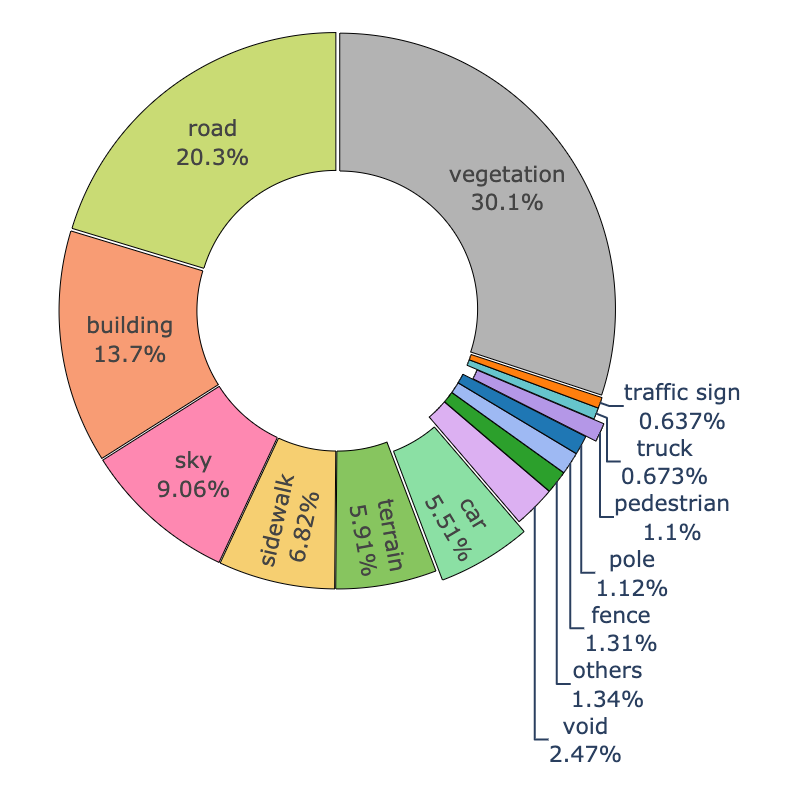} &
    \includegraphics[width=0.49\textwidth]{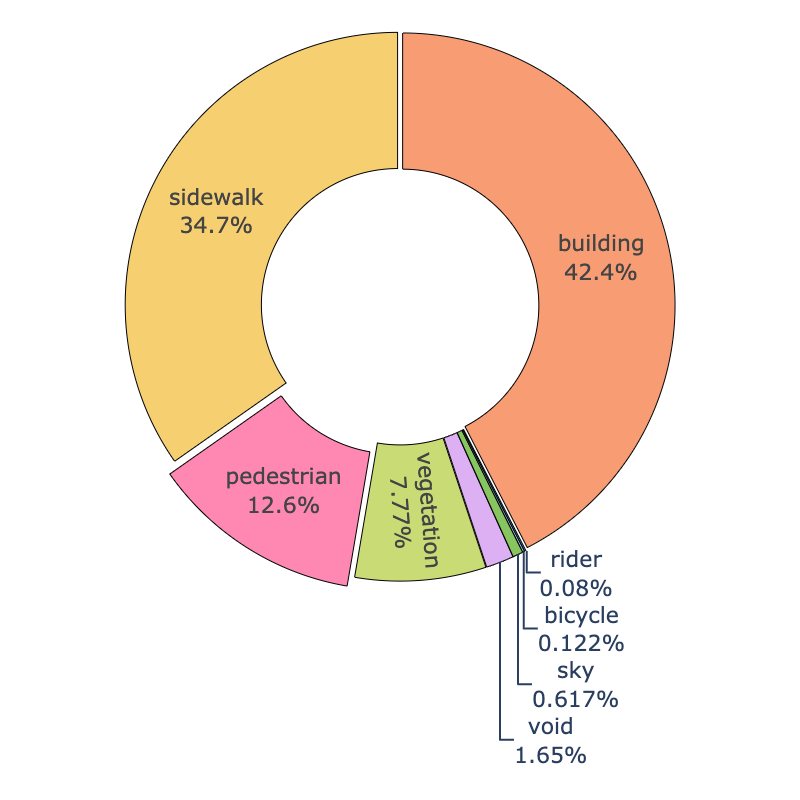} \\
    (a) \kitti. &
    (b) \challenge. \\
    \end{tabular}
    \caption{Label distribution in \kitti and \challenge.}
    \label{fig:label_dist_main}
\end{figure}

\PAR{\kitti dataset.\footnote{\url{http://cvlibs.net/datasets/kitti/eval_step.php}}} \kitti has the same train and test sequences as KITTI-MOTS (\ie, 21 and 29 sequences for training and testing, respectively). Similarly, the training sequences are further split into training set (12 sequences) and validation set (9 sequences). \changed{This dataset contains videos recorded by a camera mounted on a driving car. Since the sequence length sometimes reaches over 1000 frames, long-term tracking is required. Prior work~\cite{Valmadre2018ECCV} has shown that with increasing sequence length, the tracking difficulty increases too.
These sequences involve regularly (re-)appearing and disappearing objects, occlusions, light condition changes and scenes of pedestrian crowds.} The semantic label distribution is shown in \figref{label_dist_main}.

\PAR{\challenge dataset.\footnote{\url{https://motchallenge.net/data/STEP-ICCV21/}}} Four sequences are annotated for \challenge. In particular, we split these sequences into two for training and two for testing. This dataset contains only 7 semantic classes, as not all of Cityscapes' 19 semantic classes are present. \changed{Even though \challenge contains only one tracking class, we argue that tracking is rather difficult. First, the difficulty of tracking increases with the number of simultaneously, visually similar objects present in a scene. At the same time, this increases the chance of occlusions to occur. Concurrent work~\cite{Khurana2020Arxiv} has shown that this is indeed the case in the crowded \challenge sequences. Second, the limited number of sequences makes training harder. We pose this as an additional challenge for future work to overcome the notion of requiring large amounts of annotated data by making use of weakly-, semi- and unsupervised approaches to improve upon our baselines.} We visualize the semantic label distribution in \figref{label_dist_main}.

\begin{figure}[t]
    \begin{subfigure}[b]{0.43\linewidth}
        \centering
        \includegraphics[width=\linewidth]{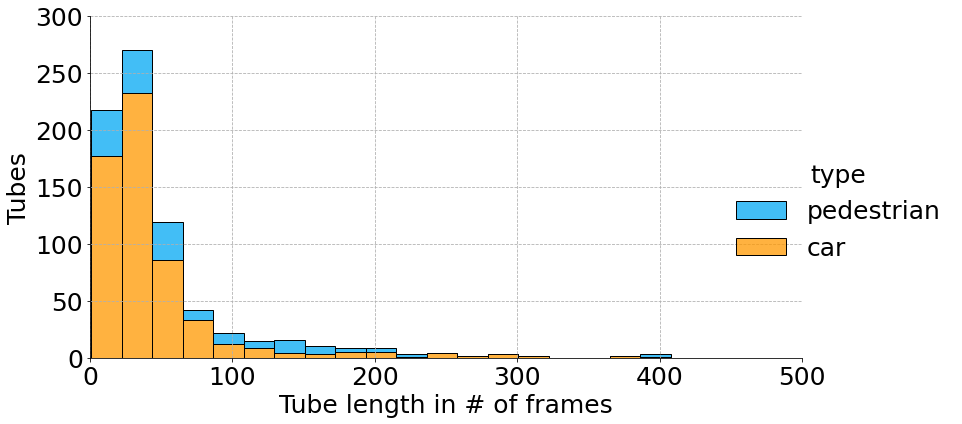}
        \caption{Track length distributions of \kitti.}
        \label{fig:tracklet_length}
    \end{subfigure} %
    \hfill
    \begin{subfigure}[b]{0.57\linewidth}
        \vspace{-8em}  
        \scalebox{0.6}{
        \begin{tabular}{l | c | c c }
            \toprule[0.2em]
            Dataset statistics & City-VPS~\cite{Kim20CVPR} & \kitti & \challenge \\
            \toprule[0.2em]
            \# Sequences (trainval/test) & 450 / 50   & 21 / 29      & 2 / 2 \\
            \# Frames (trainval/test)    & 2,700 / 300 & 8,008 / 11,095 & 1,125 / 950 \\
            \# Semantic classes          & 19       & 19         & 7 \\
            \# Annotated Masks$^\dagger$     & 72,171    & 126,529     & 17,232  \\
            Every frame annotated        & \xmark   & \cmark     & \cmark \\
            Annotated frame rate (FPS) & 3.4 & 10 & 30 \\
            \midrule
                    &  Max/Mean/Min &  Max/Mean/Min &  Max/Mean/Min \\
            Annotated frames per seq.$^\dagger$  &  6 / 6 / 6  &  1,059 / 381 / 78  & 600 / 562 / 525 \\
            Track length (frames)$^\dagger$  &   6 / 3 / 1  &  643 / 51 / 1  & 569 / 187 / 1 \\
            \bottomrule[0.1em]
        \end{tabular}
        }
        \caption{Real-world dataset comparison. $^\dagger$ refers to the trainval set.}
        \label{tab:vps_datasets}
    \end{subfigure} %
    \caption{Dataset statistics, comparison and track length distribution of \kitti.}
\end{figure}

\PAR{Dataset comparison.} In \tabref{vps_datasets}, we compare \kitti and \challenge with the only non-synthetic VPS dataset, Cityscapes-VPS~\cite{Kim20CVPR}.
Notably, we have more than 6 times the number of annotated frames and our longest video sequence is 176 times longer. 
Thus, our datasets require long-term tracking instead of focusing on short clips. \kitti contains sequences \textit{of over 1000 annotated frames} that are \textit{all} used to evaluate the performance. Hence, our datasets are not biased towards segmentation, as is the case when evaluating only on \textit{6 annotated frames}. Additionally, our pixel-dense annotations in space and time enable methods to work directly on video clips.

\PAR{Tracking Difficulty.} As shown by~\cite{Valmadre2018ECCV, Khurana2020Arxiv}, tracking difficulty primarly increases with number of objects, occlusions and tracking length. 
The track length distribution in \figref{tracklet_length} shows the importance of long-term tracking and gives evidence that evaluating on a few frames is not representative of real-world scenarios.
An extended discussing about tracking difficulty can be found in the supplement. In short, close to half of all tracks in~\cite{Kim20CVPR} are only present in a single frame requiring no tracking. 
Hence, with the available annotation budget we decided to follow the best-practices in tracking benchmarks of focusing on the most salient classes as more classes does not mean harder tracking. The remaining object classes account for less than 2\% of all pixels in the dataset.
Therefore, our datasets present a more challenging and practical scenario, requiring both short- and long-term tracking.

\PAR{Synthetic datasets.} \changed{Prior work~\cite{Kim20CVPR} has used the synthetic dataset VIPER~\cite{Richter2017ICCV}, which contains video game footage from the 2013 released game GTA5. \cite{Richter2016ECCV} has studied the realism of VIPER. Even when showing VIPER and Cityscapes images only for 500ms, the real-world data was consistently ranked to be more realistic. Moreover, \cite{Sankaranarayanan2018CVPR} has shown a large domain shift between synthetic GTA5 data and Cityscapes. Therefore, synthetic data has a purpose for pretraining, but relying purely on it does not work well enough for real-world applications, motivating the need for new benchmarks.}

%% file: sections/4.metric.tex
\section{Metric}
\label{sec:metric}
A detailed analysis of existing metrics is essential for developing a new one.
We first look into basic components PQ \cite{Kirillov19CVPR} and MOTSA \cite{Voigtlaender19CVPR_mots}, before diving into the VPQ \cite{Kim20CVPR} and PTQ \cite{Hurtado20CVPRW} metrics.
We then condense these insights into properties that a good STEP metric should satisfy.


\subsection{Metric Analysis}
\PAR{Panoptic Quality (PQ).} Kirillov \etal \cite{Kirillov19CVPR} proposed PQ to measure panoptic segmentation results.
The set of \textit{true positive} ($\tp$), \textit{false positive} ($\fp$) and \textit{false negatives} ($\fn$) segments, for a particular semantic class $c$, is formed by matching predictions $p$ to ground-truth $g$ based on the IoU scores. 
A threshold of greater than 0.5 IoU is chosen to guarantee unique matching. The overall metric is
\begin{equation}
    \label{eq:pq}
    PQ_c = \frac{\sum_{(p, g) \in \tp } IoU (p, g)}{|\tp| + \frac{1}{2} |\fp| + \frac{1}{2} |\fn|}.
\end{equation}

Defining a predicted segment with IoU of 0.51 as true positive and a segment with IoU of 0.49 as false positive (and the ground-truth segment as false negative) does not align well with human perception.
Moreover, the metric is highly sensitive to false positives that cover few pixels, which is why \textit{most, if not all,} panoptic segmentation methods use an additional post-processing step to remove {\textit stuff} classes with few pixels or add `void' class predictions~\cite{Xiong2019CVPR} to increase the PQ score.

\changed{\bm{$\Rightarrow$} A metric should be insensitive to such post-processing, and consider all predictions without requiring a threshold to define true positives on a segment level.}


\PAR{MOTA/MOTSA.} Multi-Object Tracking Accuracy (MOTA)~\cite{Bernardin08JIVP}
introduced the concept of ID switches $IDSW_c$ between frames of the prediction \wrt the ground-truth $GT_c$.
It became the standard in many tracking benchmarks \cite{Geiger12CVPR, Milan2016Arxiv}.
Its derivative Multi-Object Tracking and Segmentation Accuracy (MOTSA)~\cite{Voigtlaender19CVPR_mots} additionally considers segmentation, by matching objects based on segment overlap:
\begin{equation}
    MOTSA_c = \frac{|\tp| - |\fp| - |\idsw|}{|GT_c|}.
    \label{eq:motsa}
\end{equation}

MOTA and MOTSA were analyzed exhaustively by \cite{Luiten20IJCV}. The most relevant drawbacks are:
\setlist{nolistsep}
\begin{enumerate}[itemsep=0pt]
    \item[{D1.}] The metric penalizes ID recovery, \ie, correcting mistakes gives worse scores.
    \item[{D2.}] The metric is unbounded and can take negative values, making scores hard to interpret.
    \item[{D3.}] The tracking precision is not considered, only recall is (\cf~\cite{Luiten20IJCV} for details).
    \item[{D4.}] For detection, precision is much more important than recall, leading to a high imbalance.
    \item[{D5.}] Similar to PQ, MOTSA uses the threshold-based matching to define the set of true positives.
\end{enumerate}

\changed{\bm{$\Rightarrow$} A metric should consider both precision and recall, and should not penalize correcting mistakes.}


\PAR{VPQ.} Video Panoptic Quality (VPQ)~\cite{Kim20CVPR} for video panoptic segmentation is based on PQ and tailored to the sparsely annotated Cityscapes-VPS dataset.
VPQ computes the average quality \wrt single frames and small spans of frames, by using a \textit{2D and 3D IoU} for matching, respectively:

\begin{equation}
    \label{eq:vpq}
    VPQ = \frac{1}{K}\sum_k  \frac{1}{N_{classes}}\sum_c\frac{\sum_{(p, g) \in \tp^k } IoU_{2D/3D} (p, g)}{|\tp^k| + \frac{1}{2} |\fp^k| + \frac{1}{2} |\fn^k|}
\end{equation}
We note that these spans are constructed from every 5th video frame.
Considering this averaging and that at most $K=4$ frames are taken into account, the metric puts much more emphasis on segmentation than on association.
As noted in~\cite{Kim20CVPR}, when using more than 4 frames, the difficulty of the 3D IoU matching increases significantly. 
We agree with that assessment, but argue that we should by-pass threshold-based matching and propose a metric defined on the \emph{pixel} level. 
%
When considering a setting like ours in which there are more \textit{stuff} classes than countable \textit{things} classes, VPQ reduces this task almost to video semantic segmentation as the metric is averaged over all classes. As a consequence, the importance of association varies with the dataset class split into \textit{stuff} and \textit{things}.

\changed{\bm{$\Rightarrow$} A metric should be suitable for evaluating not only short-term but also long-term tracks as well as give equal importance to segmentation and tracking, as required for real-world scenarios.}


\PAR{(s)PTQ.} Hurtado \etal propose the Panoptic Tracking Quality (PTQ) measure \cite{Hurtado20CVPRW}:
\begin{equation}
    \label{eq:ptq}
    PTQ_c = \frac{\sum_{(p, g) \in \tp } IoU_{2D} (p, g)-|\idsw|}{|\tp| + \frac{1}{2} |\fp| + \frac{1}{2} |\fn|}.
\end{equation}
The metric combines PQ and MOTSA, by changing the
numerator
of PQ to subtract the number of ID switches $\idsw$.
The metric is computed in \textit{2D on a frame by frame basis}, thus is incapable of looking beyond \textit{first-order association}~\cite{Luiten20IJCV} mistakes.
Also, PQ's original threshold-based matching is applied.
Following this design, PTQ inherits the drawbacks from PQ as well as most issues from MOTSA as described above.

\changed{\bm{$\Rightarrow$} A metric should evaluate videos beyond single frames and decouple segmentation and association.}




\PAR{Metric Requirements.} Given these insights, we argue for a change of paradigms.
In detail, we define the following properties that a STEP metric should satisfy.
\setlist{nolistsep}
\begin{enumerate}[itemsep=0pt]
    \item [P1.] \label{item:analyze_video} {\bf Analyze full videos at pixel level:} The metric should work on \textit{full videos} and at the \textit{pixel level} rather than on single frames or at segment level.
    \item [P2.] \label{item:avoid_threshold} {\bf No threshold-based matching:} The metric should treat all pixels equally in space \textit{and} time.
    \item [P3.] \label{item:avoid_penalizing} {\bf No penalty for mistake correction:} Correcting errors in ongoing tracks should not be penalized, but encouraged to obtain long-term track consistency. 
    \item [P4.] \label{item:precision_recall} {\bf Consider precision and recall for association:}
    For association errors (\eg, ID transfer), the metric should take precision and recall into account.
    \item [P5.] \label{item:decouple_errors} {\bf Decouple errors:} 
    The metric could be decoupled into components, enabling detailed analysis for different aspects of STEP performance.
\end{enumerate}

\subsection{\metric Metric}
The previous discussions motivate us to propose the \metricfull (\metric).

\PAR{Formal Definition.} Our benchmark requires a mapping $f$ of every pixel $(x,y,t)$ (indexed by spatial coordinates $(x,y)$ and frame $t$) of a video $\Omega$ to a semantic class $c$ and a track ID $id$.
We denote the ground-truth as $gt(x, y, t)$ and the prediction as $pr(x, y, t)$.
\metric combines Association Quality (AQ) and Segmentation Quality (SQ) that measure the tracking and segmentation quality, respectively.

\PAR{Association Quality (AQ).} The proposed AQ is designed to work at a pixel-level of a full video (\cf, P1).
Importantly, all correct and incorrect associations influence the score, independent of whether a segment is above or below an IoU threshold (\cf, P2).
We define the prediction for a particular $id$ as:
\begin{align}
    pr_{id}(id) = \{(x,y,t)\ |\ pr(x,y,t) = (c, id), c \in \mathbf{C}^{th}\} \label{eq:pr_id}
\end{align}
here we only consider ``trackable`` objects $\mathbf{C}^{th}$, \ie, \textit{things}. The ground-truth is defined analogously. 
Mislabeling the semantic class of a pixel is \emph{not} penalized in AQ, and will only be judged by the \textit{segmentation quality} described later.
This decoupling prevents penalizing wrong semantic predictions twice.
We also do not require a consistent semantic class label per track, for reasons illustrated in the following example:
A ground-truth van track can easily be mistaken as a car, which should be penalized in the segmentation score, but not in the association score.
Moreover, once it becomes clear that the predicted car is actually a van, this class prediction can be corrected which results in an overall increased score for our proposed metric.
However, when requiring one semantic class per track, this prediction would be split into one track for car and one for van. As a result, this split that corrects the semantic class would receive a lower score than a prediction that keeps the wrong class, which contradicts P3.
We call this effect \textit{class-recovery} in line with the tracking term \textit{ID recovery}.

Following the notation of \cite{Luiten20IJCV}, we define the \textit{true positive associations (TPA)} of an specific ID as:
\begin{equation}
    TPA(p, g) = |pr_{id}(p) \cap gt_{id}(g)|.
\end{equation}
Similarly, \textit{false negative associations (FNA)} and \textit{false positive associations (FPA)} can be defined to compute precision $P_{id}$ and recall $R_{id}$ (\cf, P4).
Increasing precision requires the minimization of FPAs, while increasing recall requires minimization of FNAs.
We define the $IoU_{id}$ for AQ as follows:
\begin{equation}
    IoU_{id}(p, g) = \frac{P_{id}(p, g) \times R_{id}(p, g)}{P_{id}(p, g) + R_{id}(p, g) - P_{id}(p, g) \times R_{id}(p, g)}.
\end{equation}
Following our goal of long-term track consistency, we encourage \textit{ID recovery} by weighting the score of each predicted tube by their TPA.
Without this weighting, a recovered ID would not achieve a higher score.
In total, the association quality (AQ) is defined as follows.
\begin{equation}
    AQ = \frac{1}{|\mathrm{gt\_tracks}|}\sum_{g \in \mathrm{gt\_tracks}} \frac{1}{|gt_{id}(g)|}\sum_{p, |p\cap g| \neq \emptyset} TPA(p, g) \times IoU_{id}(p, g)
\end{equation}
For each ground-truth track $g$, its association score $AQ(g)$ is normalized by its size rather than by the sum of all TPA, which penalizes the removal of correct segments with wrong IDs noticeably.

\PAR{Segmentation Quality (SQ).} In semantic segmentation, Intersection-over-Union (IoU) is the most widely adopted metric~\cite{Everingham10IJCV}.
As IoU only considers the semantic labels, it fulfills the property of decoupling segmentation and association (\cf, P5).
Additionally, it allows us to measure the quality for \textit{crowd} regions\footnote{Regions of \textit{thing} classes with no distinct instances.}.
Formally, given $pr(x,y,t)$, and class $c$ we define:
\begin{align}
    pr_{sem}(c) = \{(x,y,t)\ |\ pr(x,y,t) = (c, *)\}, 
\end{align}
The ground-truth is defined analogously. We then define SQ to be the mean IoU score:
\begin{equation}
    SQ = mIoU = \frac{1}{|\mathbf{C}|}\sum_{c\in \mathbf{C}}  \frac{|pr_{sem}(c) \cap gt_{sem}(c)|}{|pr_{sem}(c) \cup gt_{sem}(c)|},
\end{equation}

\begin{figure*}
    \begin{subfigure}[t]{0.8\linewidth}
        \centering
        \includegraphics[width=\linewidth]{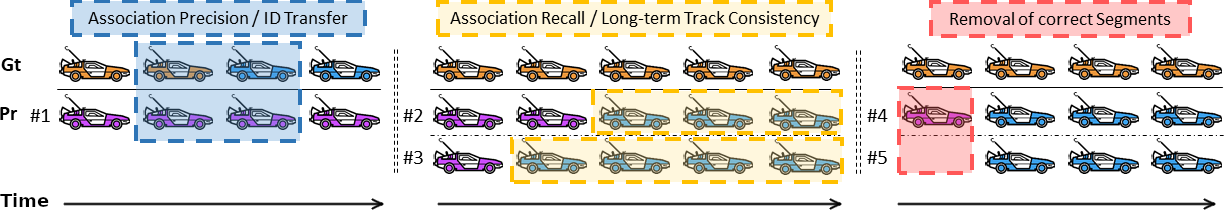}
    \end{subfigure} %
    \hfill
    \begin{subfigure}[t]{0.185\linewidth}
        \scalebox{0.6}{
            \begin{tabular}[b]{c|c c c}
                \toprule[0.2em]
                Nr. & \textbf\metric & PTQ & VPQ$^{\dagger}$\\
                \toprule[0.2em]
                \textcolor{NavyBlue}{\#1} & 0.71 & 1.0  & 0.0  \\
                \hline
                \textcolor{orange}{\#2} & 0.72 & 0.8  & 0.4  \\
                \textcolor{orange}{\#3} & 0.82 & 0.8  & 0.53 \\
                \hline
                \textcolor{OrangeRed}{\#4} & 0.79 & 0.75 & 0.5  \\
                \textcolor{OrangeRed}{\#5} & 0.65 & 0.86 & 0.75 \\
                \bottomrule[0.1em]
            \end{tabular} %
        } %
    \end{subfigure}
    \caption{An illustration of association precision, association recall and the removal of correct segments with wrong track ID for tracks of up to 5 frames. Each car is in a single-frame, where colors encode track IDs. We assume perfect segmentation and show matched tracks. For example, the left scenario contains two ground-truth tracks (orange\protect\cbox{orange}, blue\protect\cbox{RoyalBlue}), while the prediction contains a single track (violet\protect\cbox{violet}) that overlaps with both ground-truth tracks. Here, only the change of colors is important. Predictions should ideally have color transitions at the same frames as the ground-truth, if any. VPQ$^{\dagger}$ refers to the VPQ score when evaluated on full videos instead of small spans. \metric is the only metric that properly penalizes ID transfer (\#1, P4), encourages long-term track consistency (\#3 $>$ \#2, P4), and reduces the score when removing semantically correct predictions (\#4 $>$ \#5, P5). 
    }
    \vspace{5pt}
    \label{fig:metric}
\end{figure*}

\PAR{\metricfull (\metric).} The overall \metric score is the geometric mean:
\begin{equation}
    STQ = (AQ \times SQ)^{\frac{1}{2}}.
\end{equation}
%
The geometric mean is preferable over the arithmetic mean, as it rewards joint  segmentation and tracking.
Methods specializing in only one aspect of the task will therefore receive a lower score in this setting.
The effect of association errors on all proposed metrics is illustrated in \figref{metric} with intermediate computation steps in the supplement.
\tabref{metrics_comparison} provides a comparison of STEP metrics. We note that the computational complexity of all metrics is bound by the input size as the $IoU$ computation requires to read all pixels from all frames. 

\begin{table}[t]
  \centering
  \scalebox{0.8}{
  \begin{tabular}{l | c c c}
    \toprule[0.2em]
    Metric Properties & \metric & PTQ~\cite{Hurtado20CVPRW} & VPQ~\cite{Kim20CVPR} \\
    \toprule[0.2em]
    P1: Analyze full videos on pixel level & \cmark  & \xmark & (\cmark) \\
    P2: No threshold-based matching & \cmark & \xmark & \xmark \\
    P3: No penalty for mistake correction & \cmark & \xmark & \xmark \\
    P4: Consider precision and recall & \cmark & \xmark & (\cmark) \\
    P5: Decouple errors & \cmark & \xmark & \xmark \\
    \bottomrule[0.1em]
  \end{tabular}
  }
  \caption{Metric comparison. (\cmark): Partially satisfied. VPQ and PTQ fail to satisfy the properties.
  }
  \label{tab:metrics_comparison}
\end{table}




%% file: sections/5.baselines.tex
\section{Baselines}
\label{sec:baselines}
We provide single-frame and multi-frame baselines for the collected STEP datasets. Single-frame baselines follow the \textit{tracking-by-detection} paradigm, \ie, obtaining predictions in each frame independently. Then we associate predicted instances over time to obtain object tracks. Thus, we use separate modules for segmentation and tracking. By contrast, the multi-frame baseline is a {\it unified} model that jointly tackles segmentation and tracking. Architectural figures are provided in the supplement.


\PAR{Base network.}
The state-of-art panoptic segmentation model Panoptic-DeepLab~\cite{Cheng2017ICCV} is employed as the base network for both single-frame and multi-frame baselines. Panoptic-DeepLab extends DeepLab~\cite{Chen2018TPAMI,Chen2017ARXIV}, with an instance segmentation branch that predicts the instance center~\cite{Yang2019Arxiv,Zhou2019ARXIV} and regresses every pixel to its center~\cite{Ballard1981PR,Kendall2018CVPR, Weber2020IROS}. We use ResNet-50~\cite{He2016CVPR} as the network backbone.


\PAR{Single-frame baselines.}
Three different methods are used to infer the tracking IDs:
\setlist{nolistsep}
\begin{enumerate}[itemsep=0pt, leftmargin=19pt]
\item[B1.] {\bf IoU Association.}
The predicted \textit{thing} segments of two consecutive frames are matched, \ie, assigned the same tracking ID by Hungarian Matching~\cite{Kuhn1955NAVAL} with a minimal mask IoU threshold $\delta=0.3$.
To account for occluded objects, unmatched predictions are kept for $\sigma=10$ frames. Our method is insensitive to $\sigma$, \ie, using $\sigma=5, 10, 20$ yield similar results.
\item[B2.] {\bf SORT Association.} SORT~\cite{Bewley2016ICIP} is a simple online tracking method that performs bi-partite matching between sets of track predictions and object detections based on the bounding box overlap criterion. Track predictions are obtained using the Kalman filter. 
Due to its simplicity, it became a standard baseline for tasks related to tracking~\cite{dendorfer2020ijcv, Dave2020ECCV, Yang19ICCV}. In this paper, we use it as a bounding-box based instance tracking baseline using rectangles that enclose mask predictions.  
\item[B3.] {\bf Mask Propagation.}
We adopt the state-of-art optical flow method RAFT~\cite{Teed20ECCV,Sun2020ECCVW} to warp each predicted mask at frame $t-1$ into frame $t$, followed by the IoU matching (B1). We note that RAFT is highly engineered, trained on multiple datasets and achieves outstanding performance, \eg., 50\% error reduction on KITTI~\cite{Geiger12CVPR} compared to FlowNet2~\cite{Ilg2017CVPR}. We also use the forward-backward consistency check~\cite{Sundaram2010ECCV} to filter out occluded pixels during warping. 

\end{enumerate}


\PAR{Multi-frame baseline.}
Motivated by state-of-the-art MOT methods Tracktor~\cite{Bergmann19ICCV} and CenterTrack~\cite{Zhou20ECCV}, we add another prediction head, {\it previous-offset}, to the base network that regresses every pixel to its relative instance center in the {\it previous} frame. Hence, this model can predict motion directly without relying on an external network, which is {\it only possible due to our dense annotations}, \ie, such multi-frame approaches cannot be trained on sparse data as in Cityscapes-VPS.
The previous predicted center heatmap and the previous frame are given as additional inputs to the network:


\setlist{nolistsep}
\begin{enumerate}[itemsep=0pt, leftmargin=19pt]
\item[B4.] {\bf Motion-DeepLab.}
Using the predicted instance segmentation of the base network, the center of each instance is `shifted' by the {\it previous-offset} to find the closest center (within a radius $r$) in the previous frame. We apply a greedy algorithm to match instances in decreasing order of the center's score. Matched instances continue tracks, while unmatched centers start new tracks. Following~\cite{Zhou20ECCV}, $r$ equals the geometric mean of the width and height of the predicted instance. 

\end{enumerate}



%% file: sections/6.experiments.tex
\section{Results}
\label{sec:results}

We showcase the benchmark by studying the performance of different baselines on our datasets through the lens of the proposed \metric metric. 
In addition to our motion-guided baselines, we evaluate the performance of the state-of-the-art VPS model VPSNet~\cite{Kim20CVPR}. 
It uses an optical flow network~\cite{Ilg2017CVPR}, trained on external densely annotated data to align feature maps from two consecutive frames. Object instances are associated using a trained appearance-based re-id model~\cite{Yang19ICCV} among other cues. 

\PAR{Experimental Protocol.} We first pre-train all our baselines without their tracking functionality on Cityscapes~\cite{Cordts16CVPR}, and VPSNet on Cityscapes-VPS~\cite{Kim20CVPR}. These pre-trained networks are then fine-tuned on \kitti and \challenge with their tracking functionality enabled. Detailed experimental protocol is shown in the supplement. As \challenge does not have a validation set, we use sequence 9 for training and 2 for validation.


\begin{table*}[t]
  \centering
  \scalebox{0.8}{
  \begin{tabular}{l c || c c c | c | c c c c || c c c}
    \toprule[0.2em]
    \textbf{\kitti} & OF & \metric & AQ & SQ & VPQ & PTQ & sPTQ & IDS & sIDS & sMOTSA & MOTSP  \\
    \toprule[0.2em]
    B1: IoU Assoc. & \xmark & 0.58 & 0.47 & \textbf{0.71} & \textbf{0.44} & \emph{0.48} & \emph{0.48} & 1087 & 914.7 & 0.47 & \emph{0.86} \\
    B2: SORT & \xmark & \emph{0.59} & 0.50 & \textbf{0.71} & 0.42 & \emph{0.48} & \textbf{0.49} & 647 & 536.2 & 0.52 & \emph{0.86} \\
    B3: Mask Propagation & \cmark & \textbf{0.67} & \textbf{0.63} & \textbf{0.71} & \textbf{0.44} & \textbf{0.49} & \textbf{0.49} & \emph{533} & \emph{427.4} & \emph{0.54} & \emph{0.86} \\
    \hline
    B4: Motion-DeepLab & \xmark & 0.58 & 0.51 & \emph{0.67} & 0.40 & 0.45 & 0.45 & 659 & 526.7 & 0.44 & 0.84 \\
    VPSNet~\cite{Kim20CVPR} & \cmark & 0.56 & \emph{0.52} & 0.61 & \emph{0.43} & \textbf{0.49} & \textbf{0.49} & \textbf{421} & \textbf{360.0} & \textbf{0.66} & \textbf{0.91} \\
    \bottomrule[0.1em]
  \end{tabular}
  }
  \vspace{-1mm}
  \caption{We compare the different baselines under different metrics on the \kitti dataset. We highlight the \textbf{first} and \textit{second} best score in each metric. OF refers to an external optical flow network.}
  \label{tab:kitti_baselines}
\end{table*}

\begin{table*}[t]
  \centering
  \scalebox{0.8}{
  \begin{tabular}{l c || c c c | c | c c c c || c c c}
    \toprule[0.2em]
    \textbf{\challenge} & OF & \metric & AQ & SQ & VPQ & PTQ & sPTQ & IDS & sIDS & sMOTSA & MOTSP  \\
    \toprule[0.2em]
    B1: IoU Assoc. & \xmark & \textbf{0.42} & \textbf{0.25} & \textbf{0.69} & \textbf{0.55} & \textbf{0.59} & \textbf{0.59} & \emph{164} & \textbf{107.6} & 0.21 & 0.77 \\
    B2: SORT & \xmark & 0.36 & \emph{0.19} & \textbf{0.69} & \emph{0.54} & \emph{0.58} & \emph{0.58} & 364 & 254.5 & 0.22 & 0.77 \\
    B3: Mask Propagation & \cmark & \emph{0.41} & \textbf{0.25} & \textbf{0.69} & \textbf{0.55} & \emph{0.58} & \textbf{0.59} & 209 & \emph{139.4} & 0.20 & 0.77 \\
    \hline
    B4: Motion-DeepLab & \xmark & 0.35 & \emph{0.19} & \emph{0.62} & 0.51 & 0.54 & 0.54 & 326 & 227.5 & \emph{0.28}  & \emph{0.81} \\
    %
    VPSNet~\cite{Kim20CVPR} & \cmark & 0.24 & 0.17 & 0.34 & 0.25 & 0.28 & 0.28 & \textbf{152} & 146.3 & \textbf{0.40} & \textbf{0.84} \\
    \bottomrule[0.1em]
  \end{tabular}
  }
  \vspace{-1mm}
  \caption{Experimental results of different baselines on the \challenge dataset. We highlight the \textbf{first} and \textit{second} best score in each metric. OF refers to an external optical flow network.}
  \label{tab:challenge_baselines}
\end{table*}

\PAR{Experimental Results.} In \tabref{kitti_baselines}, we report results of all models on \kitti. To discuss the effects of our \metric metric empirically, we also report scores obtained with VPQ~\cite{Kim20CVPR} and PTQ~\cite{Hurtado20CVPRW}. As additional data points, we provide (soft) ID switches (s)IDS, sMOTSA and MOTS Precision (MOTSP)~\cite{Voigtlaender19CVPR_mots} scores that, in contrast to the other metrics, only evaluate `cars' and `pedestrians'. 
We observe that our single-frame baselines (\emph{B1-B3}) that perform panoptic segmentation and tracking separately achieve the overall highest \metric scores. When using multiple separate state-of-the-art models (one for panoptic segmentation and the one for optical flow), \emph{B3} achieves the highest performance in terms of association quality (AQ) and also overall in \metric. 
Both multi-frame baselines, \emph{B4} and \emph{VPSNet} are tackling a significantly more challenging problem, addressing segmentation and tracking jointly within the network. This comes at the cost of a decrease in a single-task performance (\cf SQ),
which is also observed in other multi-task settings~\cite{Kokkinos2017CVPR} such as image panoptic segmentation~\cite{Kirillov19CVPR}. Initially, separate models outperformed unified models, taking significant research effort into unified models to catch up.
\emph{B4} is such a unified model that does not rely on any external network and tackles panoptic segmentation and tracking jointly. When studying the effect of motion cues (\emph{B4}) and appearance re-id cues (\emph{VPSNet}), we notice that both significantly impact the tracking performance, as can be seen from the improvement in terms of AQ over \emph{B1}. 

From a tracking perspective, \challenge is more challenging compared to \kitti because it contains several overlapping pedestrian tracks and a reduced amount of training data. In \tabref{challenge_baselines}, we observe that \emph{B1} and \emph{B3} achieve a similar performance, which we attribute to the reduced inter-frame motion coming from pedestrians and the high 30 FPS frame rate. Notably, the \emph{SORT} tracking fails to achieve track consistency in the scenarios of many overlapping instances. Naturally, separate segmentation and tracking models (\emph{B1-B3}) are less affected by the reduced amount of training data, and therefore achieve consistently better results compared to the multi-frame baselines. However, the unified tracking and segmentation models \emph{B4} and \emph{VPSNet}, need to train their tracking heads on reduced data and therefore easily run into overfitting issues as is evident from reduced scores. Specifically, \emph{VPSNet} achieves a low SQ score, and therefore a low recall yet a high precision (\cf sMOTSA, MOTSP). In contrast, \emph{B4} does better in SQ, but cannot leverage this to improve tracking. 

In \tabref{kitti_baselines} and \tabref{challenge_baselines}, we also show the effect of different metrics. We can experimentally confirm our findings from the metric analysis (\refsec{metric}). When comparing our metric \metric with VPQ and PTQ, we observe that even though tracking quality varies significantly (\cf AQ, IDS, sMOTSA), the changes in VPQ and PTQ are rarely noticeable. This supports our theoretical findings that VPQ and PTQ evaluate this task from a segmentation perspective, leading to high scores for \emph{B1-B4} in \tabref{challenge_baselines}, and fail to accurately reflect changes in association.
This supports the need for a new perspective on this task, STEP, which equally emphasizes tracking and segmentation.

In summary, our experimental findings show that the novel \metric metric captures the essence of segmentation and tracking by considering both aspects equally.
As our datasets \kitti and \challenge provide dense annotations, for the first time  motion and appearance cues can be leveraged in a unified model for dense, pixel-precise video scene understanding. 

%% file: sections/7.conclusion.tex
\section{Conclusion}
%
%
In this paper, we present a new perspective on the task of video panoptic segmentation.
We provide a new benchmark, Segmenting and Tracking Every Pixel (STEP), to the community where we explicitly focus on measuring algorithm performance at the most detailed level possible, taking each pixel into account.
For that purpose, we analyze the drawbacks of existing metrics, and propose the \metric metric.
Our benchmark and metric are designed for evaluating algorithms in real-world scenarios, where understanding long-term tracking performance is important. 
We believe that this work provides an important STEP towards a dense, pixel-precise video understanding. 

%
\PAR{Acknowledgments.} We would like to thank S{\'e}rgio Agostinho, Deqing Sun and Siyuan Qiao for their valuable feedback, and Stefan Popov and Vittorio Ferrari for their work on the annotation tool. We would also like to thank for the supports from Google AI:D, Mobile Vision, the TUM Dynamic Vision and Learning Group, and all the Crowd Compute raters. This work has been partially funded by the German Federal Ministry of Education and Research (BMBF) under Grant No. 01IS18036B. The authors of this work take full responsibility for its content.

%% file: sections/9.supplementary.tex
\clearpage
\appendix

\section{Appendix}

In this supplementary material, we provide
\setlist{nolistsep}
\begin{enumerate}[itemsep=0pt]
    \item[B.] Benchmark checklist,
    \item[C.] An extended tracking difficulty discussion (\refsec{sup_comparison}),
    \item[D.] More of our collected dataset statistics, and details about merging our {\it semantic segmentation} annotations with existing MOTS instance annotations~\cite{Voigtlaender19CVPR_mots} (\refsec{sup_dataset}),
    \item[E.] More discussion about metric design choices (\refsec{sup_metric}),
    \item[F.] Network architecture details of our proposed unified STEP model, {\it Motion-DeepLab} (B4) (\refsec{sup_network}),
    \item[G.] More details on the experiments described in the main paper (\refsec{sup_results}),
    \item[H.] STQ on Cityscapes-VPS (\refsec{stq_vps}),
    \item[I.] Video of qualitative results: \url{https://youtu.be/NHBAvT0DXVw},
    \item[J.] Code: \url{https://github.com/google-research/deeplab2}.
\end{enumerate}

\section{Benchmark Checklist}

\begin{enumerate}
    \item[Code.] \url{https://github.com/google-research/deeplab2} includes instructions and code to reproduce our baselines.
    \item[KITTI.] \url{http://cvlibs.net/datasets/kitti/eval_step.php} includes links to the dataset, setup instructions and the test server.
    \item[MOTCh.] \url{https://motchallenge.net/data/STEP-ICCV21/} includes links to the dataset, setup instructions and the test server.
    \item[ICCV.] The proposed benchmarks are part of the 6th BMTT workshop at ICCV: \url{https://motchallenge.net/workshops/bmtt2021/}.
    \item[Resp.] The authors of this work take full responsibility for the presented work. The maintainers and creators of the original KITTI and MOTChallenge benchmarks and the MOTS datasets did consent to this project. We respect their dataset license and release the test servers in their framework.
    \item[Hosting.] The benchmarks will be hosted on the long-standing benchmark servers of KITTI and MOTChallenge. The newly provided dataset annotations will be hosted by Google. We take responsibility for the maintenance and ensure that the datasets will be accessible.
    \item[License.] We release the benchmarks with the corresponding licenses of the original datasets. The code lincense can be found at the corresponding website.
    \item[Format.] The dataset annotations are released as PNGs.
    
\end{enumerate}

\section{Tracking Difficulty Discussion}
\label{sec:sup_comparison}

\begin{figure}
    \centering
    \begin{tabular}{c}
    \includegraphics[width=0.48\linewidth]{figures/supp/mots_kitti_track_hist.png}
    \includegraphics[width=0.48\linewidth]{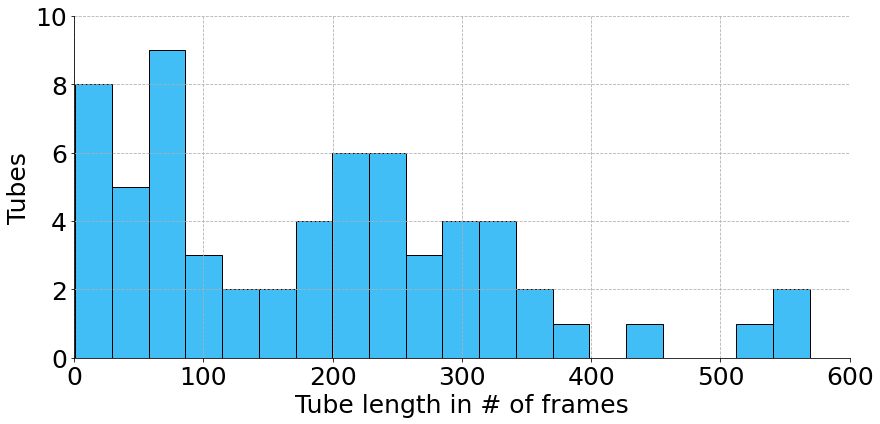}
    \end{tabular}
    \caption{Track length distributions of \kitti (left) and \challenge (right).}
    \label{fig:tracklet_length_supp}
\end{figure}

\begin{table*}
  \centering
  \scalebox{0.85}{
  \begin{tabular}{l | c c c c c c c c | c }
    \toprule[0.2em]
    Cityscapes-VPS (val) & Person & Rider & Car & Truck & Bus & Train & Motorcycle & Bicycle & All things \\
    \toprule[0.2em]
    Total instances         & 621 & 87  & 791 & 24  & 19  & 2   & 29  & 174 & 1747 \\
    Single frame instances  & 306 & 48  & 343 & 16  & 8   & 1   & 16  & 80  & 818 \\
    Spanning tracks         & 315 & 39  & 448 & 8  & 11   & 1   & 13  & 94  & 929 \\
    \hline
    Average instance length & 3.2 & 2.9 & 3.5 & 2.0 & 2.5 & 3.5 & 3.0 & 3.6 & 3.3 \\
    \bottomrule[0.1em]
  \end{tabular}
  }
  \caption{Per-class statistics of the \emph{validation} set of Cityscapes-VPS~\cite{Kim20CVPR}. Only half of all instances need to actually be tracked, \ie, appear in more than one frame. On average, instances appear for only 3 frames showing the focus on short trajectories.
  }
  \label{tab:cityscapes_per_class_statistics}
\end{table*}

\begin{table*}
    \centering
    \begin{subfigure}[t]{0.6\linewidth}
      \centering
      \begin{tabular}{l | c c | c }
        \toprule[0.2em]
        KITTI-STEP (val) & Person & Car & All things \\
        \toprule[0.2em]
        Total instances         & 151 & 68 & 219 \\
        Single frame instances  & 0 & 0 & 0 \\
        Spanning tracks         & 151 & 68 & 219 \\
        \hline
        Average instance length & 53.2 & 49.2 & 51.9 \\
        \bottomrule[0.1em]
      \end{tabular}
    \end{subfigure}
    \hfill
    \begin{subfigure}[t]{0.39\linewidth}
      \centering
      \begin{tabular}{l | c c | c }
        \toprule[0.2em]
        MOTChallenge (val) & Person \\
        \toprule[0.2em]
        Total instances         & 26 \\
        Single frame instances  & 0 \\
        Spanning tracks         & 26 \\
        \hline
        Average instance length & 183.6 \\
        \bottomrule[0.1em]
      \end{tabular}
    \end{subfigure}
    \caption{Per-class statistics of the \emph{validation} set of \kitti (left) and \challenge (right).}
    \label{tab:per_class_statistics}
\end{table*}

We provide a per-class breakdown of track numbers for the validation sets of Cityscapes-VPS~\cite{Kim20CVPR} (\tabref{cityscapes_per_class_statistics}), \kitti and \challenge (\tabref{per_class_statistics}). Furthermore, we show the track length distribution for both datasets in \figref{tracklet_length_supp}.

\PAR{Why do only `pedestrians' and `cars' have tracking IDs?}
In \figref{label_dist_main} 
in the main paper, we illustrate the class-wise histograms, \ie, amount of pixels in the whole dataset. As shown in the figure, for \kitti, both `cars' and `pedestrians' contain significantly more pixels compared to the rest of the classes, while for \challenge, the `pedestrians' dominate the others.
Due to the available annotation budget we therefore decided to focus on tracking the most salient object classes `pedestrians' and `cars' for \kitti, and only `pedestrians' for \challenge. This follows the original approach by KITTI-MOTS and MOTSChallange. We note that the number of tracking classes does \textbf{not} causally relate to the tracking difficulty. The latter is influenced by simultaneously present objects, occlusions and sequence length. The tracking difficulty and detection difficulty are changed by different factors. As our proposed benchmarks aim to balance segmentation/detection and tracking, we focus on increasing tracking difficulty \wrt previous work in this area. 


\PAR{Long-term tracking.}
In \figref{tracklet_length_supp}, we show the histograms for tracklet lengths for both datasets. As shown in the figure, our \kitti and \challenge present a challenge for long term consistency in segmentation and tracking.

\PAR{Do more instances or classes lead to harder tracking?} Even though, our datasets provide twice the number of masks for the trainval set compared to Cityscapes-VPS, Cityscapes-VPS provides significantly more unique \emph{instances}. Here, instances refer to unique objects not distinguishing for how many frames they are visible. We refrain from calling them tracks, as tracks are usually implied to last more than a single frame. When comparing these numbers directly, \tabref{cityscapes_per_class_statistics} shows that the validation set provides 1747 unique instances, while \kitti and \challenge (\tabref{per_class_statistics}) provide only 219 and 26 instances. In the following we provide several reasons why more instances does not imply harder tracking:

\begin{enumerate}
    \item In Cityscapes-VPS, almost half (818) of all instances (1747) only last for a single frame, hence requiring no tracking at all. This shows the strong focus of Cityscapes-VPS on the segmentation aspect.
    \item The average length of instances is 3.3 on Cityscapes-VPS, \ie, instances need to be tracked for 2.3 frames after being detected. Hence, Cityscapes-VPS is not suitable to measure tracking performance. On \kitti and \challenge, the average length are 51.9 and 183.6 frames, respectively.
    \item The validation set of Cityscapes-VPS consists of 50 clips while our datasets contain 10 videos in the validation set. Naturally, every new clip and video always introduces a new set of instances. However, in long videos, instances continue to exist throughout (part of) the video resulting in less new instances. Yet, exactly these instances are the ones that need to be tracked and the ones that make tracking challenging.
\end{enumerate}

We therefore infer that the proposed datasets \kitti and \challenge are significantly more suitable when evaluating segmentation and tracking than Cityscapes-VPS. Measuring tracking requires (long) trajectories in the data.

\section{Extended Dataset Discussion}
\label{sec:sup_dataset}

\PAR{Merging annotation.} We merge our new semantic segmentation annotation with the existing tracking instance ground-truth, \ie, instance identity from the KITTI-MOTS and MOTS-Challenge~\cite{Voigtlaender19CVPR_mots}. 
We refer to their annotations as MOTS annotations for simplicity. 
\figref{dataset_main} from the main paper gives an example of our annotation process. 
During the merging process, potential inconsistencies between our annotated semantic labels for the classes `pedestrian' and `car' and the MOTS instance annotations need to be carefully handled.
For example, following the `pedestrian' definition in Cityscapes~\cite{Cordts16CVPR}, our semantic annotation includes items carried (but not touching the ground) by the person, while MOTS annotations exclude those items.
With the aim to build a dataset that is compatible for both Cityscapes and MOTS definitions, we adopt the following strategy to merge the semantic annotations.
For the cases where our annotated semantic mask is larger than MOTS annotated mask, we dilate the MOTS instance annotations with a kernel size of 15. The difference between the intersection of our annotation with the the enlarged and with the original MOTS annotations is re-labeled as ‘VOID’.
In practice, the union regions are detected by dilating the MOTS instance annotations with a kernel size 15.
For the cases where our annotated semantic mask is smaller than MOTS annotated mask, the inconsistent regions are overwritten with MOTS annotations.
As a result, the consistent `pedestrian' masks are annotated with both semantic class and tracking IDs. Small regions along the masks are annotated with `void', while personal items are annotated with only the semantic class (and tracking ID 0). 
Additionally, the ignored regions in the MOTS annotations are filled with our semantic annotation. For the crowd regions, \eg, a group of `pedestrians' that could not be distinguished by annotators,  semantic labels are annotated but their tracking ID is set to 0. 
More technically, we denote each MOTS instance mask with semantic label \textit{l} and instance ID \textit{i} as \(M_l^i\), and the semantic annotation mask with label \textit{k} as \(S_k\). We first dilate \(M_l^i\) using a kernel with 15 pixels. There are three cases for each \(S_k\):

\textbf{Case I}: \(S_k\) intersects with \(M_l^i\). The intersection is overwritten with label \textit{l} with instance ID = \textit{i}.

\textbf{Case II}: \(S_k\) intersects with the expanded dilated region. The intersection is re-labeled as `void' if \(k = l\).

\textbf{Case III}: The semantic label of the rest without any intersection remains the same.

\begin{figure}
    \centering
    \includegraphics[width=0.45\textwidth]{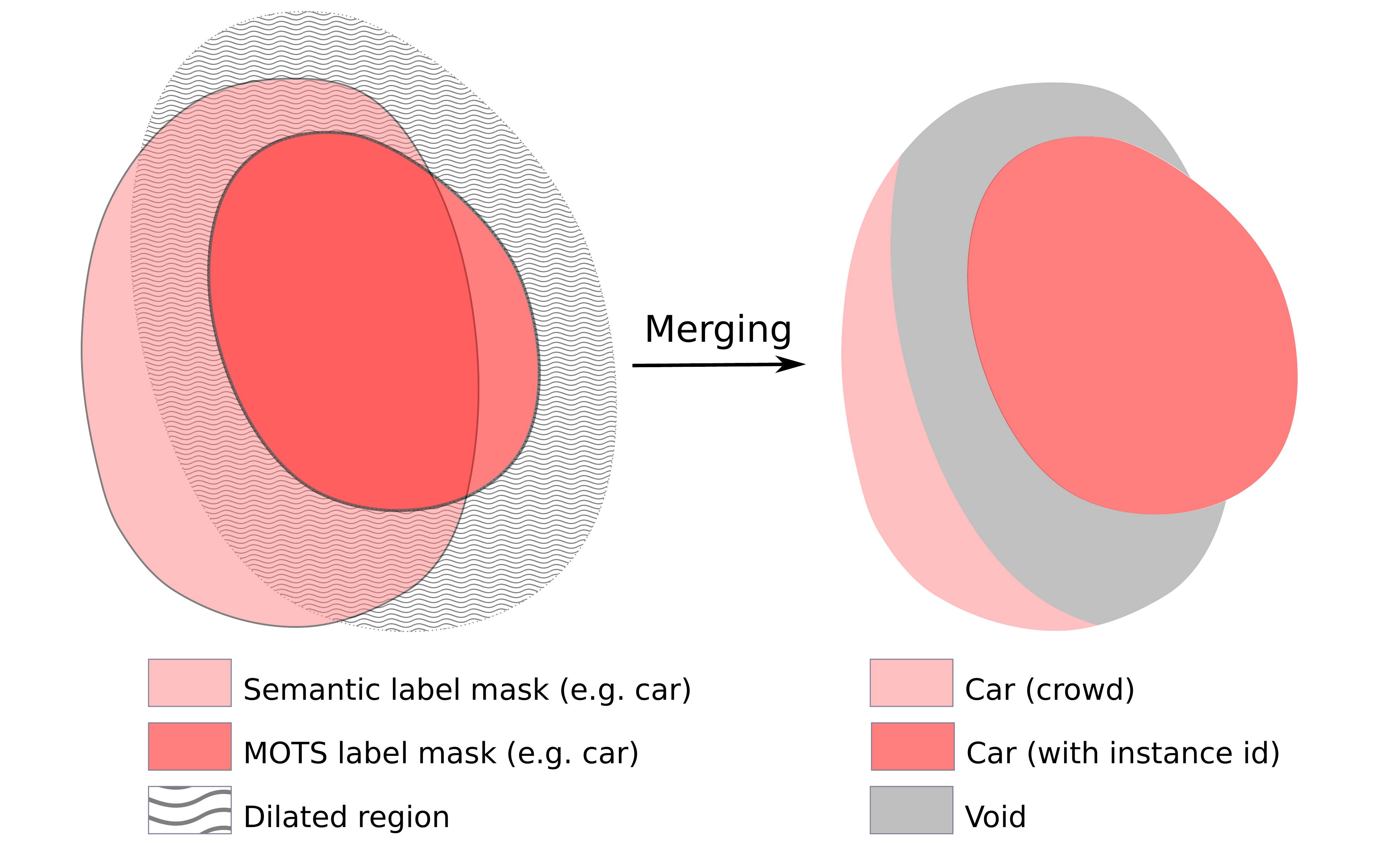}
    \caption{Illustration of how we merge the semantic and MOTS annotations. A large dilation region is chosen for illustration purpose.}
    \label{fig:merging}
\end{figure}

This process is summarized in \figref{merging}. We refer to the merged datasets as \kitti and \challenge, respectively.

\section{Extended Metric Discussion}
\label{sec:sup_metric}
In addition to the overview of metrics in the main paper, we discuss one more common metric in video instance segmentation~\cite{Yang19ICCV}, and explain why it is unsuitable for STEP. After that, we will discuss our design choices of \metricfull (\metric).

\subsection{Track-mAP (${AP}^{track}$)}

For the task of Video Instance Segmentation~\cite{Yang19ICCV}, a variant of ${AP}^{mask}$~\cite{Hariharan2014ECCV,Lin14ECCV} is used to measure the quality of predictions. Like ${AP}^{mask}$, ${AP}^{track}$ allows overlapping predictions, and hence requires confidence scores to rank instance proposals. These predictions are then matched with an IoU threshold.
Moreover, as established in prior work \cite{Luiten20IJCV}, this metric can be gamed by making lots of low-confidence predictions, and the removal of correct detections with wrong track ID can improve scores.
We therefore consider this metric unsuitable for our benchmark.

\subsection{\metric Design Choices}

As stated in the main paper, we define the association quality (AQ) as follows:

\begin{align}
    AQ(g) &= \frac{1}{|gt_{id}(g)|}\sum_{p, |p\cap g| \neq \emptyset} TPA(p, g) \times IoU_{id}(p, g),\nonumber \\
    AQ &= \frac{1}{|\mathrm{gt\_tracks}|}\sum_{g \in \mathrm{gt\_tracks}} AQ(g).
    \label{eq:aq}
\end{align}

\PAR{Precision and Recall.} While it is common for recognition and segmentation tasks to consider both recall and precision, this has not been widely adapted for tracking yet. For example, MOTSA does not consider precision for association. However, precision is important to penalize predicted associations that are false, \ie, \textit{false positive associations}. Consider the following example: All cars in a sequences are segmented perfectly and are assigned the same track ID. As all ground-truth pixels are covered, this gives perfect recall. Yet, the overall prediction is far from being perfect by assigning the same track ID to different cars. Hence, precision is an important aspect of measuring the quality of a prediction.
The other aspect to consider is recall. Considering the same example with perfect segmentation, a perfect association precision can be trivially achieved by assigning a different track ID to every pixel. As there are no false positives associations, the overall score is perfect. Yet, this does not fulfill the purpose of measuring the quality of association. Therefore, both aspects, precision and recall, have to be considered for a good metric measuring association.

\PAR{IoU \vs F1.} The two most common approaches to combine precision and recall in computer vision are Intersection-over-Union, also known as the Jaccard Index, and F1, also known as the dice coefficient. IoU and F1 correlate positively and are thus both valid measures. We chose IoU for two reasons:
\begin{enumerate}
    \item We already adopted the IoU metric for measuring segmentation. Using it for association as well leads to a more consistent formulation.
    \item When comparing F1 and the IoU score, the F1 score is the harmonic mean of precision and recall and therefore closer to the average of both terms. On the other hand, IoU is somewhat closer to the minimum of precision and recall. Choosing IoU over F1 therefore emphasizes that good predictions need to consider recall and precision for association as well as highlighting innovation better.
\end{enumerate}

\PAR{Weighting factor \emph{TPA}.} A simpler version of equation \eqref{eq:aq} would compute the average $IoU_{id}$ score without any weighting, and by normalizing with the number of partially overlapping predictions \wrt the ground-truth track. However, this formulation has the disadvantage that it does not consider long-term consistency of each track. Given two predicted tracks A and B, whether the IoU to the single ground-truth track are $3/5$ and $2/5$ or $4/5$ and $1/5$, both would achieve the exact same result:

\begin{equation}
    \frac{1}{2} \times \left(\frac{3}{5}+\frac{2}{5}\right) = \frac{1}{2} \times \left(\frac{4}{5}+\frac{1}{5}\right) = \frac{1}{2}
\end{equation}

As our goal is long-term consistency, we weight each $IoU_{id}$ with the $TPA$. This factor increases the importance of long-term prediction:

\begin{align}
    \frac{1}{5} \times \left(3 \times \frac{3}{5} + 2 \times \frac{2}{5}\right) &= \frac{13}{25}\\
    \frac{1}{5} \times \left(4 \times \frac{4}{5} + 1 \times \frac{1}{5}\right) &= \frac{17}{25}
\end{align}

Thus, our formulation of AQ fulfills the property of getting a higher score for predictions that have overall higher long-term consistency.

\PAR{Normalization by ground-truth size.} When considering the normalization factor of equation \eqref{eq:aq}, one natural question that could come up is, why do we propose this denominator instead of the sum of all \emph{TPA}. The reason is that otherwise the removal of correctly segmented regions with wrong track ID could achieve a higher score. Consider two predicted car tracks overlapping a ground-truth car track with IoU $4/5$ and $1/5$, respectively. Changing the denominator would lead to the following scores, with and and without the removal of the second track:

\begin{align}
    \frac{1}{5} \times \left(4 \times \frac{4}{5} + 1 \times \frac{1}{5}\right) &= \frac{17}{25}\\
    \frac{1}{4} \times \left(4 \times \frac{4}{5}\right) &= \frac{16}{20} = \frac{20}{25}
\end{align}

Hence, the removed segment leads to a higher score. In contrast, in our current formulation we achieve the following scores in this scenario:

\begin{align}
    \frac{1}{5} \times \left(4 \times \frac{4}{5} + 1 \times \frac{1}{5}\right) &= \frac{17}{25}\\
    \frac{1}{5} \times \left(4 \times \frac{4}{5}\right) &= \frac{16}{25}
\end{align}

Therefore, it will always be better to recognize cars (or other objects) than not to detect them. This still holds when looking at the overall metric. Even though the removal of correct segments is already penalized in the segmentation quality, that penalty would would rarely be noticeable when the association quality score would increase in that case. Hence, setting the denominator to the ground-truth size aligns with the importance of not removing predictions. For example, in an autonomous driving scenario, it is critical that correct pedestrian predictions are kept, even though they have a wrong track ID.

\PAR{Class-aware \vs Class-agnostic Association.} In previous metrics, VPQ~\cite{Kim20CVPR} and PTQ~\cite{Hurtado20CVPRW} tracks must have the correct semantic class assigned to count as \emph{true positives}. Such design couples segmentation and association errors, \eg, a car track mistaken for a van would receive a score of 0 even though is is perfectly tracked throughout a sequence. In our setting, we compare three options to design the association score \wrt to semantic classes.

\begin{enumerate}
    \item Require the \emph{correct} semantic class of predicted tracks to be matched to ground-truth tracks to compute association sores.
    \item Require \emph{one} (but any) semantic class of predicted tracks to be matched to ground-truth tracks to compute association sores.
    \item Allow \emph{any} semantic thing class to be assigned to pixels of predicted tracks.
\end{enumerate}

Option 1 penalizes wrong semantic segmentation twice and therefore completely couples segmentation and association errors like VPQ and PTQ. The 2nd option has the problem that correcting semantic classes receives a lower score than not correcting them. A prediction that at first mistakes a van for a car should not be penalized, when the semantic class is changed to the correct one. When requiring one semantic class, a prediction that changes the semantic class would create a new track. This would result in an overall reduced score, which violates the goal of not penalizing the correction of mistakes. Therefore, we have chosen the 3rd option for the design of our \metric metric.

\PAR{Implementation Details.} We need to consider two special cases for the implementation of the \metric metric. The first case is the \emph{crowd} region. For far away or highly overlapping objects, it can be impossible for human annotators to distinguish different instances. In those cases, we can still assign the correct semantic class to these pixels. During evaluation, we cannot measure any association quality, but also do not want to penalize (potentially correct) track ID assignment by a network. We therefore consider the semantic class of these regions for measuring the segmentation quality. For the association quality, these pixel regions are ignored, which means there is no penalty for assigning track IDs to these regions.
The second case to consider is the \emph{ignore} label. Ignore labels are commonly used by annotators for regions, which can not be assigned to one of the limited semantic classes and should therefore be ignored during evaluation. However, \cite{Kirillov19CVPR} introduced this concept for predictions, too. Predicted ignore segments do not count as false positives, which lead to common post-processing steps in the field of panoptic segmentation. Specifically, small predicted segments are overwritten with the void label.
Since we would not like to encourage such tricks, we adopt the following strategy to handle void label.
For the segmentation quality, we allow an additional class void, which is handled like all other classes, except that all ignore regions in the ground-truth will not be considered. Thus, there is no advantage of predicting void labels, but we still allow to evaluate output of methods that require such predictions by design.

\begin{table*}
  \centering
  \scalebox{0.75}{
    \begin{tabular}{c | lllll}
    \toprule[0.2em]
        & \#1 & \#2 & \#3 & \#4 & \#5 \\
    \toprule[0.2em]
    SQ & 1.0 & 1.0 & 1.0 & 1.0 & 0.75 \\
    AQ & $\frac{1}{2\times2}(\frac{2\times2}{4}+\frac{2\times2}{4})=0.5$ & $\frac{1}{5}(\frac{2\times2}{5}+\frac{3\times3}{5})=\frac{13}{25}$ & $\frac{1}{5}(\frac{1\times1}{5}+\frac{4\times4}{5})=\frac{17}{25}$ & $\frac{1}{4}(\frac{1\times1}{4}+\frac{3\times3}{4})=\frac{5}{8}$ & $\frac{1}{4}(\frac{3\times3}{4})=\frac{9}{16}$\\
    STQ & $\sqrt{1\times0.5}=0.71$ & $\sqrt{1\times\frac{13}{25}}=0.72$ &  $\sqrt{1\times\frac{17}{25}}=0.82$ & $\sqrt{1\times\frac{5}{8}}=0.79$ & $\sqrt{\frac{3}{4}\times\frac{9}{16}}=0.65$ \\
    \hline\\
    PTQ & $\frac{4-0}{4+0+0}=1.0$ & $\frac{5-1}{5+0+0}=0.8$ & $\frac{5-1}{5+0+0}=0.8$ & $\frac{4-1}{4+0+0}=0.75$ & $\frac{3-0}{3+\frac{1}{2}+0}=0.86$\\ \\
    \hline\\
    VPQ$^\dagger$ & $\frac{0}{0+\frac{1}{2}+2\times\frac{1}{2}}=0$ & $\frac{0.6}{1+\frac{1}{2}+0}=0.4$ & $\frac{0.8}{1+\frac{1}{2}+0}=0.53$ & $\frac{0.75}{1+\frac{1}{2}+0}=0.5$ & $\frac{0.75}{1+0+0}=0.75$\\ \\
    \bottomrule[0.1em]
    \end{tabular}
    }
    \caption{Intermediate computation steps for Fig. 3 of the main paper. VPQ$^\dagger$ refers to the VPQ evaluation over the complete scene.}
    \label{tab:intermediate_steps}
\end{table*}

\PAR{Detailed metric scores of illustration.} We provide intermediate computation steps to obtain the results of Fig. 3 of the main paper in \tabref{intermediate_steps}.

\section{Network Architecture}
\label{sec:sup_network}

\PAR{Single-frame baselines.} Our single-frame baselines build on top of Panoptic-DeepLab~\cite{Cheng20CVPR} by additionally using three different methods to infer the tracking IDs. The adopted separate architectures are therefore the same as the original works~\cite{Cheng20CVPR,Teed20ECCV}.

\begin{figure*}
    \centering
    \includegraphics[width=1.00\textwidth]{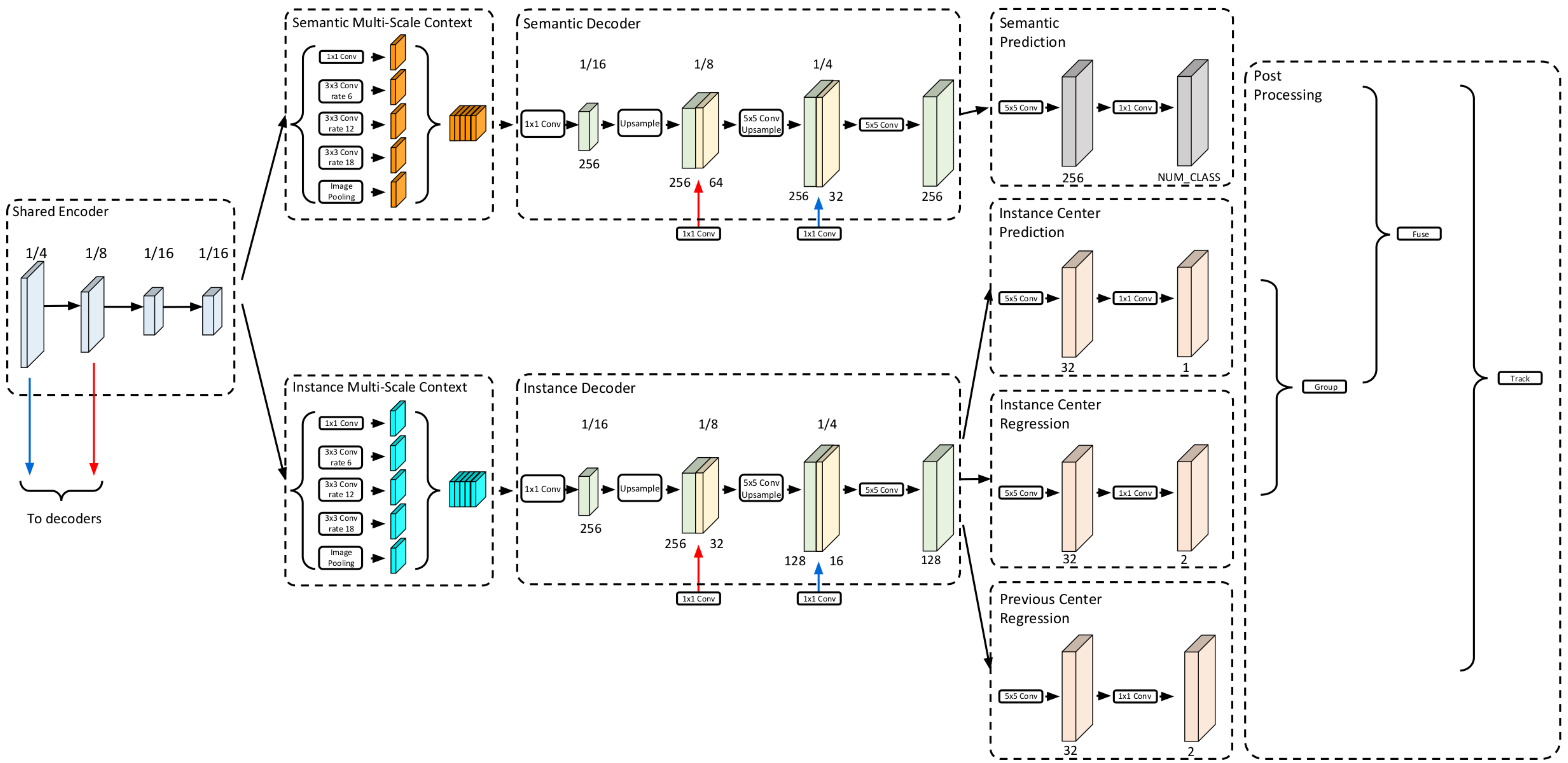}
    \caption{The architecture of the \emph{Motion-DeepLab} baseline (B4).}
    \label{fig:network_arch}
\end{figure*}

\PAR{Multi-frame baseline.} Motivated by~\cite{Bergmann19ICCV,Zhou20ECCV}, our multi-frame baseline, \emph{`Motion-DeepLab'}, extends Panoptic-DeepLab~\cite{Cheng20CVPR} by adding another prediction head, \textit{Previous Center Regression}, which assists in associating predicted instances between two consecutive frames. Additionally, same as CenterTrack~\cite{Zhou20ECCV}, the previous predicted center heatmap and the previous image frame are given as additional inputs to the network. The network architecture is visualized in \figref{network_arch}.

\section{Experimental Results}
\label{sec:sup_results}

\PAR{Training Protocol.}
Using Panotic-DeepLab~\cite{Chen2020CVPR} as our base network, we follow closely the same training protocol as in~\cite{Chen2020CVPR}.
Specifically, all our models are trained using TensorFlow~\cite{Abadi2016tUSENIX} on 16 TPUs and batch size 32. We use the `poly' learning rate policy~\cite{Liu2015ARXIV}, fine-tune the batch normalization parameters~\cite{Ioffe2015ICML}, adopt random scale data augmentation during training with Adam~\cite{Kingma2015ICLR} optimizer. Our model is pretrained on Cityscapes~\cite{Cordts16CVPR} (only image panoptic annotations are exploited) for 60k iterations with an initial learning rate of 2.5e-4. For the single-frame baselines (B1-B3), we fine-tune on \kitti and \challenge with an initial learning rate of 1e-5 for 30k and 1.4k iterations, respectively.
For our \emph{Motion-DeepLab} baseline (B4), we have to conduct net-surgery on the weights of the first convolution of the ResNet pre-trained checkpoint~\cite{He2016CVPR}. The baseline B4 takes 7 channels as input (3 channels for current frame, 3 channels for previous frame, and 1 channel for the previous frame center heatmap). We therefore take the weights of the first $7\times7$ convolution and duplicate them to get to 6 channels. Finally, the weights of the last channel are obtained by taking another duplicate and average over the channel dimension. With these pre-trained weights, we fine-tune on \kitti and \challenge with a learning rate of 1e-5 for 50k and 2k iterations, respectively.
For VPSNet~\cite{Kim20CVPR}, we use the default training settings to pre-train on Cityscapes-VPS without the tracking head. Then, we fine-tune the full network on \kitti and \challenge again with the optimized default settings. As we observe overfitting on \challenge, we reduce the training iterations to $1/3$ of the original number.

\PAR{Qualitative Results.} Please refer to the attached files for video visualization of our dataset ground truth and our model predictions (B3).

\PAR{Effect of pre-training.} In \tabref{ablation_kitti}, we report the effect of pretraining our networks on Cityscapes before finetuning on \kitti. As shown in the table, pretraining brings 10\%, and 4\% improvement of STQ for baselines B3 and B4, respectively. For B4, we observe performance gain in SQ, and slightly degradation in AQ, presenting a challenging research problem to efficiently develop a unified STEP model. When comparing the non-pretrained networks, the unified model B4 has a much smaller gap to the B3 model than with pretraining. We hope our baseline could serve as a strong baseline to facilitate the research along the direction of developing a better unified STEP model.

\begin{table*}
  \centering
  \begin{tabular}{l c | c c c || c c c }
    \toprule[0.2em]
    Baseline & Pretrained & \metric & AQ & SQ & PQ & RQ & SQ \\
    \toprule[0.2em]
    B3: Mask Propagation & \xmark & 0.57 & 0.59 & 0.55 & 0.36 & 0.46 & 0.72 \\
    B3: Mask Propagation & \cmark & 0.67 & 0.63 & 0.71 & 0.47 & 0.57 & 0.79 \\
    \hline
    B4: Motion-DeepLab & \xmark & 0.54 & 0.55 & 0.53 & 0.34 & 0.43 & 0.69 \\
    B4: Motion-DeepLab & \cmark & 0.58 & 0.51 & 0.67 & 0.43 & 0.54 & 0.78 \\
    \bottomrule[0.1em]
  \end{tabular}
  \caption{{\bf Effect of pretraining} on Cityscapes with results on \kitti.
  }
  \label{tab:ablation_kitti}
\end{table*}

\section{STQ on Cityscapes VPS}
\label{sec:stq_vps}

In \tabref{vps_on_cityscapes_per_class} and \tabref{vps_on_cityscapes}, we show scores of our metric with  VPSNet on Cityscapes-VPS. In the following, we draw insights from these numbers.

\PAR{Metric insights on Cityscapes-VPS.} In \tabref{vps_on_cityscapes}, we provide \metric scores of VPSNet on Cityscapes-VPS. Notably, the AQ score is lower than on \kitti but higher than on \challenge. In the following, we study the behavior of AQ. For that we also provide per-class scores on Cityscapes-VPS in \tabref{vps_on_cityscapes_per_class}.

\begin{table*}
  \centering
  \begin{tabular}{l | c c c || c c c }
    \toprule[0.2em]
    Cityscapes-VPS (val) & \metric & AQ & SQ & VPQ & VPQ\textsuperscript{Th} & VPQ\textsuperscript{St} \\
    \toprule[0.2em]
    VPSNet & 0.50 & 0.35 & 0.72 & 0.57 & 0.44 & 0.67 \\
    \bottomrule[0.1em]
  \end{tabular}
  \caption{VPSNet evaluated on Cityscapes-VPS~\cite{Kim20CVPR}. Scores obtained from official code and models.
  }
 \label{tab:vps_on_cityscapes}
\end{table*}


\begin{table*}
  \centering
  \scalebox{0.8}{
  \begin{tabular}{l | c c c c c c c c | c | c}
    \toprule[0.2em]
    Cityscapes-VPS (val) & Person & Rider & Car & Truck & Bus & Train & Motorcycle & Bicycle & All things & All \\
    \toprule[0.2em]
    VPQ ($K=4$) & 0.32 & 0.35 & 0.45 & 0.32 & 0.34 & 0.43 & 0.27 & 0.28 & 0.34 & 0.52\\
    \hline
    STQ       & 0.51 & 0.48 & 0.63 & 0.46 & 0.38 & 0.31 & 0.26 & 0.44 & 0.50 & 0.50 \\
    AQ        & 0.30 & 0.30 & 0.42 & 0.29 & 0.20 & 0.18 & 0.18 & 0.27 & 0.35 & 0.35 \\
    \bottomrule[0.1em]
  \end{tabular}
  }
  \caption{Per-class breakdown of thing scores of VPSNet evaluated on the validation set of Cityscapes-VPS~\cite{Kim20CVPR}. Scores obtained from official code and models. VPQ with $K=4$ annotated frames, is the longest and hardest evaluation of VPQ under the official settings.
  }
  \label{tab:vps_on_cityscapes_per_class}
\end{table*}

Due to our unified treatment of space and time, \metric can evaluate single frame performance, short clip performance as well as long video performance. Depending on the dataset, the AQ score will therefore adapt to the given characteristics. Naturally, AQ cannot measure tracking when the dataset focus is on segmentation as discussed above. Hence, AQ will measure pixel-precise image and small clip instance segmentation on Cityscapes-VPS. A lower AQ on Cityscapes-VPS does therefore not imply that tracking is harder than on \kitti. Moreover, all instances contribute equally towards the final score independent of the instance size. With almost 50\% instances not being tracks, 50\% of the score will be image instance segmentation. 

As a reminder, VPQ is averaged over the scores when evaluating PQ on one, two, three, and four frames. In this example, AQ behaves similar as VPQK when evaluated on $K=4$ frames.